\documentclass[final]{cvpr}

\usepackage{times}
\usepackage{epsfig}
\usepackage{graphicx}
\usepackage{amsmath}
\usepackage{amssymb}

\usepackage{multirow}
\usepackage{booktabs}
\usepackage{subcaption}

\usepackage{color}

\usepackage[pagebackref=true,breaklinks=true,colorlinks,bookmarks=false]{hyperref}

\begin{document}

%%%%%%%%% TITLE
\title{Contrastive Embedding for Generalized Zero-Shot Learning}

\newcommand*\samethanks[1][\value{footnote}]{\footnotemark[#1]}
\author{%
	Zongyan Han\textsuperscript{1}\samethanks[2] , ~~%
	Zhenyong Fu\textsuperscript{1}\thanks{Corresponding authors.}~\thanks{Zongyan
		Han, Zhenyong Fu and Jian Yang are with PCA Lab, Key Lab of
		Intelligent Perception and Systems for High-Dimensional
		Information of Ministry of Education, and Jiangsu Key Lab of Image
		and Video Understanding for Social Security, School of Computer
		Science and Engineering, Nanjing University of Science and
		Technology, China.} , ~~%
	Shuo Chen\textsuperscript{2,1}~%\thanks{Shuo Chen is with RIKEN Center for Advanced Intelligence Project, Japan.And he is also with School of Computer Science and Engineering, Nanjing University of Science and Technology, China.}~~%
	and Jian Yang\textsuperscript{1}\samethanks[1]  \samethanks[2]%
	\\
	% Institutes
	{\textsuperscript{1} PCALab, Nanjing University of Science and Technology, China} \\
	{\textsuperscript{2} RIKEN Center for Advanced Intelligence Project, Japan} \\
	{\tt\small {\{hanzy, z.fu, csjyang\}@njust.edu.cn}}~~
	{\tt\small shuo.chen.ya@riken.jp} \\
	\\
	% Emails
	
}

%\newcommand*\samethanks[1][\value{footnote}]{\footnotemark[#1]}
%\author{Zongyan Han\textsuperscript{1}\samethanks[2] 
%	\and Zhenyong Fu\textsuperscript{1}\thanks{Corresponding authors.} \thanks{Zongyan
%		Han, Zhenyong Fu and Jian Yang are with PCA Lab, Key Lab of
%		Intelligent Perception and Systems for High-Dimensional
%		Information of Ministry of Education, and Jiangsu Key Lab of Image
%		and Video Understanding for Social Security, School of Computer
%		Science and Engineering, Nanjing University of Science and
%		Technology, China.}
%	\and Shuo Chen\textsuperscript{2,1}\thanks{Shuo Chen is with RIKEN Center for Advanced Intelligence Project, Japan.
%		And he is also with School of Computer
%		Science and Engineering, Nanjing University of Science and
%		Technology, China.}
%	\and Jian Yang\textsuperscript{1}\samethanks[1]  \samethanks[2]
%\\
%	{\textsuperscript{1} instXXX} ~
%{\textsuperscript{2} instYYY} ~
%	{\tt\small mailXXX, ~~mailYYY} \\
%}

%\date{%
%	$^1$Organization 1\\%
%	$^2$Organization 2\\[2ex]%
%%	\today
%}

\maketitle
%\pagestyle{empty}
%\thispagestyle{empty}

%%%%%%%%% ABSTRACT
\begin{abstract}  
Generalized zero-shot learning (GZSL) aims to recognize objects from
both seen and unseen classes, when only the labeled examples from seen 
classes are provided. Recent feature generation methods learn a
generative model that can synthesize the missing visual features of
unseen classes to mitigate the data-imbalance problem in
GZSL. However, the original visual feature space is suboptimal for
GZSL classification since it lacks discriminative information. To
tackle this issue, we propose to integrate the generation model with
the embedding model, yielding a hybrid GZSL framework. The hybrid GZSL
approach maps both the real and the synthetic samples produced by the
generation model into an embedding space, where we perform the final
GZSL classification. Specifically, we propose a contrastive embedding
(CE) for our hybrid GZSL framework. The proposed contrastive embedding
can leverage not only the class-wise supervision but also the
instance-wise supervision, where the latter is usually neglected by
existing GZSL researches. We evaluate our proposed hybrid GZSL
framework with contrastive embedding, named CE-GZSL, on five benchmark
datasets. The results show that our CE-GZSL method can outperform the
state-of-the-arts by a significant margin on three datasets.
Our codes are available
on~\url{https://github.com/Hanzy1996/CE-GZSL}.
\end{abstract}

%-------------------------------------------------------
\section{Introduction}
Object recognition is a core problem in computer vision.
This problem on a fixed set of categories with plenty of training
samples has progressed tremendously due to the advent of deep 
convolutional neural networks~\cite{krizhevsky2012imagenet}.
However, realistic object categories often follow a long-tail
distribution, where some categories have abundant training samples and
the others have few or even no training samples available.
Recognizing the long-tail distributed object categories is
challenging, mainly because of the imbalanced training sets of these
categories.
Zero-Shot Learning (ZSL)~\cite{lampert2009learning,palatucci2009zero}
holds the promise of tackling the extreme data imbalance between
categories, thus showing the potential of addressing the long-tail
object recognition problem.
Zero-shot learning aims to classify objects from previously unseen
categories without requiring the access to data from those
categories.
In ZSL, a recognition model is first learned on the seen categories,
of which the training samples are provided.
Relying on the category-level semantic descriptors, such as visual
attributes~\cite{farhadi2009describing,lampert2009learning} or word
vectors~\cite{mikolov2013efficient,mikolov2013distributed}, ZSL
can transfer the recognition model from seen to unseen object
categories in a data-free manner.

\begin{figure}[t]
  \centering
  \includegraphics[width=0.85\linewidth]{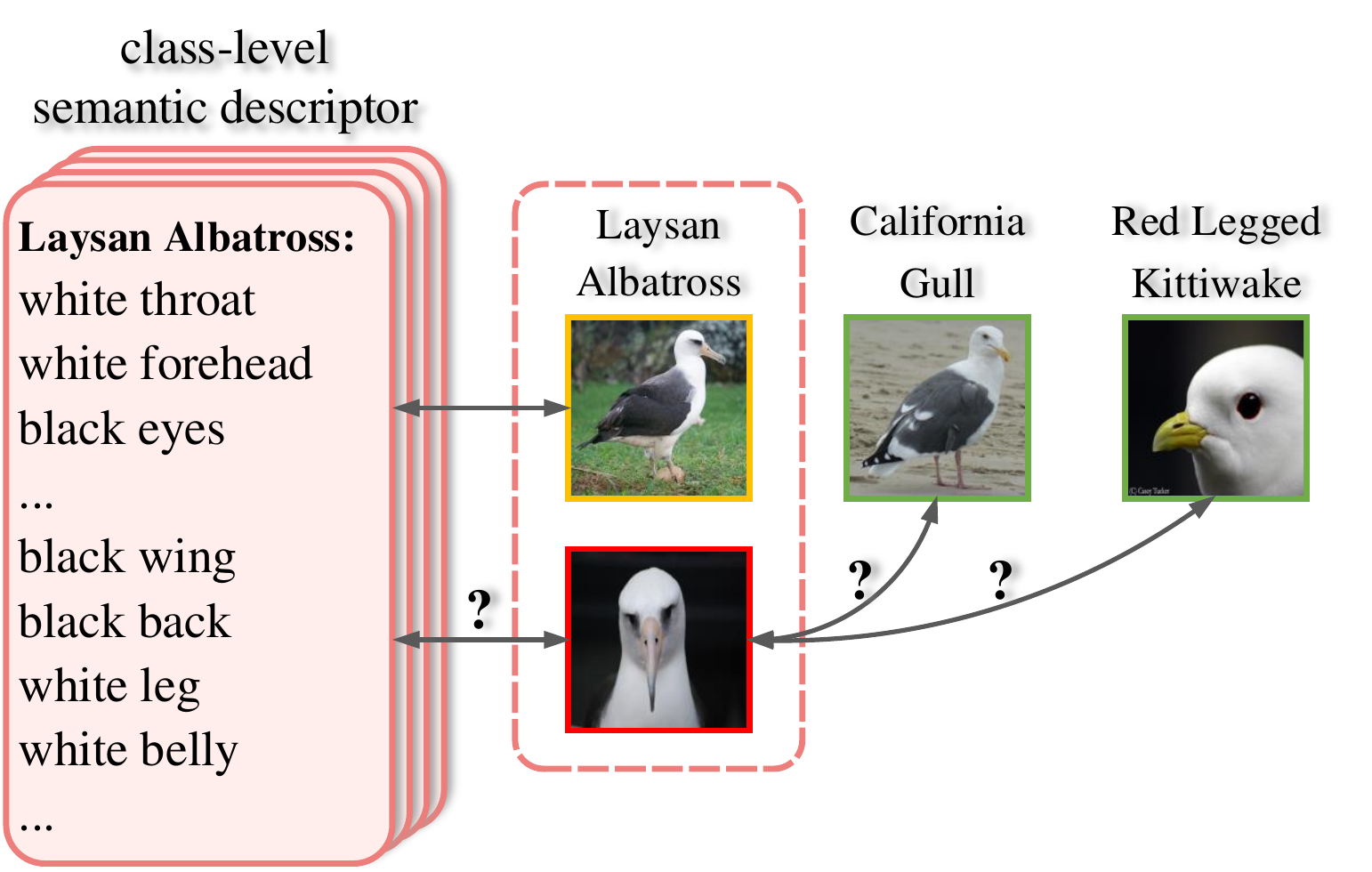}
  \caption{Existing semantic embedding methods merely utilize the
    class-wise supervision, which may be unsuitable for some examples
    as they do not match exactly with the class-level semantic
    descriptor. The proposed contrastive embedding can utilize not
    only the class-wise supervision but also the instance-wise
    supervision.}
  \label{fig:motivation}
\end{figure}

In zero-shot learning, we have the available data from seen classes
for training. Conventional zero-shot
learning~\cite{akata2013label,socher2013zero} assumes that the test
set contains the samples from unseen classes only, while in the recent
proposed Generalized Zero-Shot Learning
(GZSL)~\cite{chao2016empirical,xian2017zero}, the test set is composed
of the test samples from both seen and unseen classes. A large body of
conventional ZSL methods learns a semantic embedding function to map
the visual features into the semantic descriptor
space~\cite{frome2013devise,akata2015evaluation,romera2015embarrassingly,zhang2015zero,fu2017zero}. In
the semantic space, we can conduct the ZSL classification by directly
comparing the embedded data points with the given class-level semantic
descriptors. Semantic embedding methods excel in conventional ZSL, yet
their performance degrades substantially in the more challenging GZSL
scenario, owing to their serious bias towards seen classes in the
testing phase~\cite{xian2018zero}. Conventional ZSL is unnecessary to
worry about the bias problem towards seen classes as they are excluded
from the testing phase. But in GZSL the bias towards seen classes will
make the GZSL model misclassify the testing images from unseen
classes.

To mitigate the bias problem in GZSL, feature generation based GZSL
methods have been
proposed~\cite{bucher2017generating,mishra2018generative,kumar2018generalized,xian2018feature,xian2019f,shen2020invertible}
to synthesize the training samples for unseen classes.
The feature generation method can compensate for the lack of training
samples of unseen classes.
Merging the real seen training features and the synthetic unseen
features yields a fully-observed training set for both seen and unseen
classes.
Then we can train a supervised model, such as a softmax classifier, to
implement the GZSL classification.
However, the feature generation methods produce the synthesized visual
features in the \emph{original} feature space.
We conjecture that the original feature space, far from the semantic
information and thus lack of discriminative ability, is suboptimal for
GZSL classification.

To get the best of both worlds, in this paper, we propose a hybrid
GZSL framework, grafting an embedding model on top of a feature
generation model. In our framework, we map both the real seen features
and the synthetic unseen features produced by the feature generation
model to a new embedding space. We perform the GZSL classification in
the new embedding space, but not in the original feature space.

Instead of adopting the commonly-used semantic embedding
model~\cite{frome2013devise,akata2015evaluation}, we propose a
\emph{contrastive embedding} in our hybrid GZSL framework. The
traditional semantic embedding in ZSL relies on a ranking loss, which
requires the correct (positive) semantic descriptor to be ranked
higher than any of wrong (negative) descriptors with respect to the
embedding of a training sample. The semantic embedding methods only
utilize the class-wise supervision. In contrastive embedding, we wish
to exploit not only the class-wise supervision but also the
instance-wise supervision for GZSL, as depicted in
Figure~\ref{fig:motivation}. Our proposed contrastive embedding learns
to discriminate between one positive sample (or semantic descriptor)
and a large number of negative samples (or semantic descriptors) from
different classes by leveraging the contrastive
loss~\cite{gutmann2010noise,oord2018representation,wu2018unsupervised}. We
evaluate our method on five benchmark datasets, and to the best of our
knowledge, our method can outperform the state-of-the-arts on three
datasets by a large margin and achieve competitive results on the
other two datasets.

Our contributions are three-fold:
(1) we propose a hybrid GZSL framework combining the embedding based
model and the feature generation based model;
(2) we propose a contrastive embedding, which can utilize both the
class-wise supervision and the instance-wise supervision, in our
hybrid GZSL framework;
and (3) we evaluate our GZSL model on five benchmarks and our method
can achieve the state-of-the-arts or competitive results on these
datasets.

%-------------------------------------------------------
\section{Related Work}
Zero-shot learning~\cite{lampert2009learning,palatucci2009zero} aims
to transfer the object recognition model from seen to unseen classes
via the shared semantic space, in which both seen and unseen classes
have their semantic descriptors.
Early ZSL works focus on the conventional ZSL problem. These works
typically learn to embed visual samples and the semantic descriptors
to an embedding
space~\cite{frome2013devise,akata2013label,fu2014transductive,akata2015evaluation,fu2015zero,kodirov2015unsupervised,fu2015transductive,romera2015embarrassingly,bucher2016improving,kodirov2017semantic,cacheux2019modeling}
(e.g. the visual space or the semantic descriptor space).
In the embedding space, the visual samples from the same class are
supposed to center around the corresponding class-level semantic
descriptor. They implement conventional ZSL recognition by searching
the nearest semantic descriptor in the embedding space.
In the more challenging GZSL scenario, however, embedding-based
methods suffer from the seen classes overfitting problem due to the
data-imbalance nature of ZSL~\cite{xian2017zero}.
To relieve the overfitting problem, some
methods~\cite{chao2016empirical,liu2018generalized,annadani2018preserving,huynh2020fine,xie2019attentive,xie2020region,min2020domain}
have designed new loss functions to balance the predictions between
seen and unseen classes. Some other works~\cite{mandal2019out,
  keshari2020generalized, chen2020boundary} have regarded GZSL as an
out-of-distribution detection problem. Moreover, some
researches~\cite{lee2018multi,wang2018zero,liu2020hyperbolic} have
introduced the knowledge graph in GZSL to propagate the learned
knowledge from seen to unseen classes through the knowledge graph.

To further mitigate the data imbalance problem, feature generation methods
learn to complement the visual samples for unseen
classes~\cite{bucher2017generating,mishra2018generative,kumar2018generalized,xian2018feature,xian2019f,paul2019semantically,sariyildiz2019gradient,vyas2020leveraging}. The
feature generation methods first learn a conditional generative model
based on such as Variational Autoencoder (VAE)~\cite{kingma2013auto}
and Generative Adversarial Networks
(GAN)~\cite{goodfellow2014generative,arjovsky2017wasserstein},
conditioned on the semantic descriptors.
With the learned generative model, they can synthesize the missing
visual examples for unseen classes using the corresponding semantic
descriptors. With the real examples from seen classes and the
synthesized examples from unseen classes, they can transform the GZSL
problem into a standard supervised classification problem and learn a
supervised classifier to implement GZSL recognition. Recently, Shen
\etal~\cite{shen2020invertible} have introduced Generative
Flows~\cite{dinh2014nice,dinh2016density,kingma2018glow} into
zero-shot learning and achieved good performance for GZSL and
conventional ZSL.

Though existing methods have achieved great success on GZSL, as
discussed before, the \emph{original} visual feature space lacks the
discriminative ability and is suboptimal for GZSL classification.
Therefore, we propose a hybrid GZSL framework, integrating a feature
generation model with an embedding based model. Inspired by the
emerging contrastive representation
learning~\cite{gutmann2010noise,oord2018representation,wu2018unsupervised,he2020momentum,khosla2020supervised}, 
we propose a contrastive embedding model for our hybrid GZSL
framework, in which we consider both the instance-wise supervision and
the class-wise supervision. In contrast, the traditional semantic
embedding for ZSL only utilizes the class-wise supervision. Our hybrid
GZSL framework maps the real seen samples and the synthetic unseen
samples into a new embedding space, where we learn a supervised
classifier, e.g. softmax, as the final GZSL
classifier.

%-------------------------------------------------------
\section{Contrastive Embedding for GZSL}
\label{sec:cont_embedding_gzsl}
In this section, we first define the Generalized Zero-Shot Learning
(GZSL) problem, before introducing the proposed hybrid GZSL framework
and the contrastive embedding in it.

\subsection{Problem definition}
In ZSL, we have two disjoint sets of classes: $S$ seen classes in
$\mathcal{Y}_s$ and $U$ unseen classes in $\mathcal{Y}_u$, where we
have $\mathcal{Y}_{s}\cap\mathcal{Y}_{u}=\varnothing$.
Suppose that $N$ labeled instances from seen classes $\mathcal{Y}_s$
are provided for training:
$\mathcal{D}_{tr}=\{(x_1,y_2),\dots,(x_N,y_N)\}$, where
$x_i\in\mathcal{X}$ denotes the instance and $y_i\in\mathcal{Y}_s$ is
the corresponding seen class label.
The test set $\mathcal{D}_{te}=\{x_{N+1},\dots,x_{N+M}\}$ contains $M$
unlabeled instances.
In conventional ZSL, the instances in $\mathcal{D}_{te}$ come from
unseen classes only. Under the more challenging Generalized Zero-Shot
Learning (GZSL) setting, the instances in $\mathcal{D}_{te}$ come from
both seen and unseen classes. At the same time, the class-level
semantic descriptors of both seen and unseen classes are also provided
$\mathcal{A}=\{a_1,\dots,a_{S},a_{S+1},\dots,a_{S+U}\}$, where the
first $S$ semantic descriptors correspond to seen classes in
$\mathcal{Y}_{s}$ and the last $U$ semantic descriptors correspond to
unseen classes in $\mathcal{Y}_{u}$. We can infer the semantic
descriptor $a$ for a labeled instance $x$ from its class label $y$.

\begin{figure*}[t]
  \centering
  \includegraphics[width=0.98\linewidth]{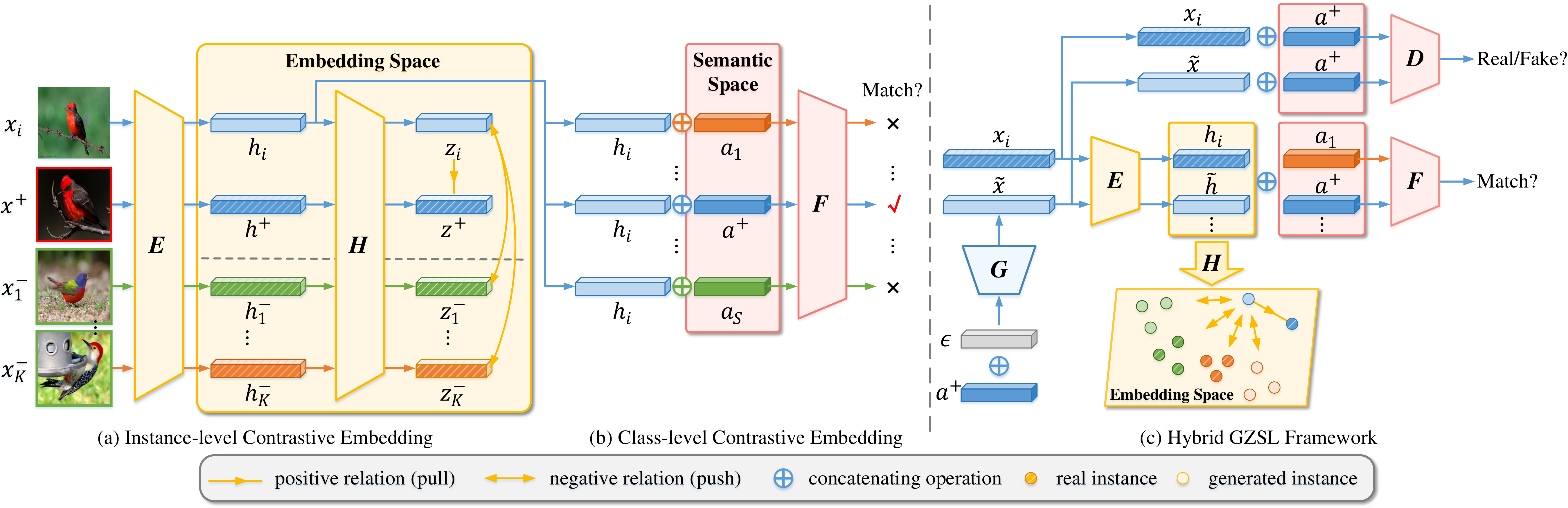}
  \caption{Illustration of our proposed hybrid GZSL framework with
    contrastive embedding (CE-GZSL). We learn an embedding function
    $E$ that maps the visual samples $x_i$ into the embedding space as
    $h_i=E(x_i)$. 
    We further learn a non-linear projection $H$ to better constrain the  embedding space: $z_i=H(h_i)$.
    We introduce a comparator network $F$ that measures
    the relevance score between $h_i$ and the semantic descriptors. We
    learn the embedding function with both the instance-level and the
    class-level supervisions. We integrate the contrastive embedding
    model with the feature generation model. In the feature generation
    model, the feature generator $G$ learns to produce visual features
    based on a semantic descriptor $a$ and a Gaussian noise
    $\epsilon$; and the discriminator $D$ aims to distinguish the fake
    visual features from real ones.}
  \label{fig:all_structure}
\end{figure*}

%-------------------------------------------------------
\subsection{A Hybrid GZSL Framework}
\label{sec:a_hybrid_gzsl}
Semantic embedding (SE) in conventional ZSL aims to learn an embedding
function $E$ that maps a visual feature $x$ into the semantic
descriptor space denoted as $E(x)$.
The commonly-used semantic embedding methods rely on a structured
loss function proposed in~\cite{akata2015evaluation,frome2013devise}.
The structured loss requires the embedding of $x$ being closer to the
semantic descriptor $a$ of its ground-truth class than the descriptors
of other classes, according to the dot-product similarity in the
semantic descriptor space.
Concretely, the structured loss is formulated as below:
\begin{small}
\begin{equation}
  \mathcal{L}_{se}^{real}(E)=\mathbb{E}_{p(x,a)}[\max(0,\Delta-a^{\top}E(x)+(a^\prime)^{\top}E(x))],
  \label{eq:L_se_real}
\end{equation}
\end{small}where $p(x,a)$ is the empirical distribution of the real
training samples of seen classes, $a^\prime\neq a$ is a
randomly-selected semantic descriptor of other classes, and $\Delta>0$
is a margin parameter to make $E$ more robust.

Semantic embedding methods are less effective in GZSL due to the
severe bias towards seen classes. Recently, many feature generation
methods~\cite{xian2018feature,kumar2018generalized,mishra2018generative,huang2019generative,atzmon2019adaptive}
have been proposed to synthesize the missing training samples for
unseen classes. Feature generation methods learn a conditional
generator network $G$ to produce the samples $\tilde{x}=G(a,\epsilon)$
conditioned on a Gaussian noise
$\epsilon\sim\mathcal{N}(\mathbf{0},\mathbf{I})$ and a semantic
descriptor $a$. In the meanwhile, a discriminator network $D$ is
learned together with $G$ to discriminate a real pair $(x,a)$ from a
synthetic pair $(\tilde{x},a)$. The feature generator $G$ tries to
fool the discriminator $D$ by producing indistinguishable synthetic
features. The feature generation methods hope to match the synthetic
feature distribution with the real feature distribution in the
original feature space. The feature generator network $G$ and the
discriminator network $D$ can be learned by optimizing the following
adversarial objective:
\begin{small}
\begin{equation}
  \begin{aligned}
    V(G,D)=&\mathbb{E}_{p(x,a)}[\log D(x,a)]\\
    &+\mathbb{E}_{p_{G}(\tilde{x},a)}[\log (1-D(\tilde{x},a))],
  \end{aligned}
  \label{eq:loss_adv}
\end{equation}
\end{small}where $p_{G}(\tilde{x},a)=p_{G}(\tilde{x}|a)p(a)$ is the
joint distribution of a synthetic feature and its corresponding
semantic descriptor.

The feature generation methods learn to synthesize the visual features
in the \emph{original} feature space. However, in the original feature
space, the visual features are usually not well-structured and thus
are suboptimal for GZSL classification. In this paper, we propose a
hybrid GZSL framework, integrating the embedding model and the feature
generation model. In our hybrid GZSL framework, we map both the real
features and the synthetic features into an embedding space, where we
perform the final GZSL classification. In its simplest form, we just
choose the semantic descriptor space as the embedding space and
combine the learning objective of semantic embedding defined in
Eq.~\ref{eq:L_se_real} and the objective of feature generation defined
in Eq.~\ref{eq:loss_adv}. To map the synthesized features into the
embedding space as well, we introduce the following embedding loss for
the synthetic features:
\begin{small}
  \begin{equation}
    \begin{aligned}
      \mathcal{L}_{se}^{sync}(G,E)=\mathbb{E}_a[\max(0,\Delta&-a^{\top}E(G(a,\epsilon))\\
      &+(a^\prime)^{\top}E(G(a,\epsilon)))].
    \end{aligned}
    \label{eq:L_se_sync}
  \end{equation}
\end{small}Notably, we formulate $\mathcal{L}_{se}^{sync}(G,E)$ only
using the semantic descriptors of seen classes. Therefore, the total
loss of our basic hybrid GZSL approach takes the form of
%\begin{small}
\begin{equation}
  \begin{aligned}
    \max_{D}\min_{G,E}V(G,D)+\mathcal{L}_{se}^{real}(E)+\mathcal{L}_{se}^{sync}(G,E).
  \end{aligned}
\label{eq:se_gen_loss}
\end{equation}
%\end{small}

%-------------------------------------------------------
\subsection{Contrastive Embedding}
\label{sec:contrastive_embedding}
Our basic hybrid GZSL framework is based on the traditional semantic
embedding model, where only the class-wise supervision is
exploited. In this section, we present a new contrastive embedding
(CE) model for our hybrid GZSL framework. The contrastive embedding
consists of the instance-level contrastive embedding based on the
instance-wise supervision and the class-level contrastive embedding
based on the class-wise supervision.

\paragraph{Instance-level contrastive embedding}
In the embedding space, the embedding of a visual sample $x$ is
denoted as $h=E(x)$. For each data point $h_i$ embedded from either a
real or synthetic seen feature, we set up a $(K+1)$-way classification
subproblem to distinguish the unique one positive example $h^+$ from
total $K$ negative examples $\{h_1^-,\dots,h_K^-\}$. The positive
example $h^+$ being randomly selected has the same class label with
$h_i$, while the class labels of the negative examples are different
from $h_i$'s class label. Here, we follow the strategy
in~\cite{chen2020simple} to add a non-linear projection head $H$ in
the embedding space: $z_i=H(h_i)=H(E(x_i))$. And we perform the
$(K+1)$-way classification on $z_i$ to learn the embedding
$h_i$. Concretely, the cross-entropy loss of this $(K+1)$-way
classification problem is calculated as follows:
\begin{small}
  \begin{equation}
    \label{eq:l_ce_ins}
    \ell_{ce}^{ins}(z_i,z^+)
    =-\log\frac{\exp{\left(z_i^{\top}z^+/\tau_e\right)}}{\exp{\left(z_i^{\top}z^+/\tau_e\right)}+
      \sum_{k=1}^K\exp{\left(z_i^{\top}z_k^-/\tau_e\right)}},
  \end{equation}
\end{small}where $\tau_e>0$ is the temperature parameter for the
instance-level contrastive embedding and $K$ is the number of negative
examples. Intuitively, a large $K$ will make the problem in
Eq.~\ref{eq:l_ce_ins} more difficult. The large number of negative
examples encourages the embedding function $E$ to capture the strong
discriminative information and structures shared by the samples, real 
and synthetic, from the same class in the embedding space.

To learn the embedding function $E$, the non-linear projection $H$ and
the feature generator network $G$, we calculate the loss function for
the instance-level contrastive embedding as the expected loss computed
over the randomly selected pairs $z_i$ and $z^{+}$ for both the real
and synthetic examples, where $z_i\neq z^+$ but they belong to the
same seen class.
\begin{equation}
  \label{eq:L_ce_ins}
  \mathcal{L}_{ce}^{ins}(G,E,H)=\mathbb{E}_{z_i,z^+}\left[\ell_{ce}^{ins}(z_i,z^+)\right].
\end{equation}

\paragraph{Class-level contrastive embedding}
Analogously, we can formulate a class-level contrastive
embedding. Since we do not limit our embedding space to be the
semantic descriptor space, we cannot compute the dot-product
similarity between an embedded data point and a semantic descriptor
directly. Thus, we learn a comparator network $F(h,a)$ that measures
the relevance score between an embedding $h$ and a semantic descriptor
$a$. With the help of the comparator network $F$, we formulate the
class-level contrastive embedding loss for a randomly selected point
$h_i$ in the embedding space as an $S$-way classification subproblem.
The goal of this subproblem is to select the only one correct semantic
descriptor from total $S$ semantic descriptors of seen classes. In
this problem, the only positive semantic descriptor is the one
corresponding to $h_i$'s class, while the remaining $S-1$ semantic
descriptors from the other classes are treated as the negative
semantic descriptors. Similarly, we can calculate the cross-entropy
loss of this $S$-way classification problem as below:
\begin{small}
  \begin{equation}
    \label{eq:l_ce_cls}    
    \ell_{ce}^{cls}(h_i,a^+)
    =-\log\frac{\exp{\left(F(h_i,a^+)/\tau_s\right)}}{\sum_{s=1}^S\exp{\left(F(h_i,a_s)/\tau_s\right)}},
  \end{equation}
\end{small}where $\tau_s>0$ is the temperature parameter for the
class-level contrastive embedding and $S$ is the number of seen
classes. The class-level contrastive embedding relies on the
class-wise supervision to strengthen the discriminative ability of the
samples in the new embedding space.

We define the following loss function for the class-level contrastive
embedding:
\begin{equation}
  \label{eq:L_ce_cls}
  \mathcal{L}_{ce}^{cls}(G,E,F)=\mathbb{E}_{h_i,a^+}\left[\ell_{ce}^{cls}(h_i,a^+)\right],
\end{equation}which is the expected loss over the samples, either real
or synthetic, in the new embedding space, and their corresponding
semantic descriptor, i.e. the positive descriptor.

%-------------------------------------------------------
% \vspace{-0.2cm}
\paragraph{Total loss}
In our final hybrid GZSL framework, we replace the semantic embedding
(SE) model in the basic hybrid framework in Eq.~\ref{eq:se_gen_loss}
with the proposed contrastive embedding (CE) model.
As described above, the contrastive embedding model consists of an
instance-level loss function $\mathcal{L}_{ce}^{ins}$ and a
class-level loss function $\mathcal{L}_{ce}^{cls}$.
Thus, the total loss of our final hybrid GZSL framework with
contrastive embedding (CE-GZSL) is formulated as:
% \begin{small}
\begin{equation}
  \begin{aligned}
    \max_D\min_{G, E,H, F}V(G,D)+\mathcal{L}_{ce}^{ins}(G,E,H)+\mathcal{L}_{ce}^{cls}(G,E,F).
  \end{aligned}
  \label{eq:ce_gen_loss}
\end{equation}Figure~\ref{fig:all_structure} illustrates the whole
structure of our method. In our method, we learn a feature generator
$G$ (together with a discriminator $D$) to synthesize the missing
unseen class features; we learn an embedding function $E$ to embed the
samples, both real and synthetic, to a new embedding space, where we
conduct the final GZSL classification; to learn a more effective
embedding space, we introduce a non-linear projection $H$ in the
embedding space which is used to define the instance-level contrastive
embedding loss; and to enforce the class-wise supervision, we learn a
comparator network $F$ to compare an embedding and a semantic
descriptor.

% \vspace{-0.2cm}
\paragraph{GZSL classification}
We first generate the features for each unseen class in the embedding
space by composing the feature generator network $G$ and the embedding
function $E$: $\tilde{h}_j=E(G(a_u,\epsilon))$, where $u\geq S+1$ and
$a_u$ is the semantic descriptor of an unseen class. We map the given
training features of seen classes in $\mathcal{D}_{tr}$ into the same
embedding space as well: $h_i=E(x_i)$. In the end, we utilize the real 
seen samples and the synthetic unseen samples in the embedding space
to train a softmax model as the final GZSL classifier.

%-------------------------------------------------------
\section{Experiments}
\paragraph{Datasets}
We evaluate our method on five benchmark datasets for ZSL: Animals
with Attributes 1\&2 (\textbf{AWA1}~\cite{lampert2009learning} \&
\textbf{AWA2}~\cite{xian2018zero}), Caltech-UCSD Birds-200-2011
(\textbf{CUB})~\cite{wah2011caltech}, Oxford Flowers
(\textbf{FLO})~\cite{nilsback2008automated}, and SUN Attribute (\textbf{SUN})~\cite{patterson2012sun}.
AWA1 and AWA2 share the
same 50 categories and each category is annotated with 85 attributes,
which we use as the class-level semantic descriptors. AWA1 contains
30,475 images and AWA2 contains 37,322 images; CUB contains 11,788
images from 200 bird species; FLO contains 8,189 images of 102
fine-grained flower classes; SUN contains 14,340 images from 717
different scenes and each class is annotated with 102 attributes. For
the semantic descriptors of CUB and FLO, we adopt the 1024-dimensional
class embeddings generated from textual
descriptions~\cite{reed2016learning}. We extract the 2,048-dimensional
CNN features for all datasets with ResNet-101~\cite{he2016deep}
pre-trained on ImageNet-1K~\cite{krizhevsky2012imagenet} \emph{without
  finetuning}.
Moreover, we adopt the Proposed Split (PS)~\cite{xian2018zero} to
divide all classes on each dataset into seen and unseen classes.

\paragraph{Evaluation Protocols}
We follow the evaluation strategy proposed in
~\cite{xian2018zero}. Under the conventional ZSL scenario, we only
evaluate the per-class Top-1 accuracy on unseen classes. Under the
GZSL scenario, we evaluate the Top-1 accuracy on seen classes and
unseen classes, respectively, denoted as ${S}$ and ${U}$. The
performance of GZSL is measured by their harmonic mean:
${H}=2\times{S}\times{U}/({S}+{U})$.

\begin{table*}[ht!]
  \caption{Comparisons with the state-of-the-art GZSL methods. $U$ and
    $S$ are the Top-1 accuracies tested on unseen classes and seen
    classes, respectively, in GZSL. $H$ is the harmonic mean of $U$
    and $S$. The best results are marked in bold.}
  \vspace{-0.1cm}
  \centering
  \resizebox{0.9\textwidth}{!}
  {
    \label{table:comp_sota}
    \begin{tabular}{l|ccc|ccc|ccc|ccc|ccc}
      \toprule
      \multirow{2}*{Method}&\multicolumn{3}{|c|}{AWA1}&\multicolumn{3}{c|}{AWA2}&\multicolumn{3}{c|}{CUB}&\multicolumn{3}{c}{FLO}&\multicolumn{3}{|c}{SUN}\\
                           &${U}$&${S}$&${H}$&${U}$&${S}$&${H}$&${U}$&${S}$&${H}$&${U}$&${S}$&${H}$&${U}$&${S}$&${H}$\\
      \midrule
      DAZLE~\cite{huynh2020fine}&-&-&-&60.3&75.7&67.1&\underline{56.7}&59.6&58.1&-&-&-&\underline{52.3}&24.3&33.2\\
      TCN~\cite{jiang2019transferable}&49.4&76.5&60.0&\underline{61.2}&65.8&63.4&52.6&52.0&52.3&-&-&-&31.2&37.3&34.0\\
      Li \etal~\cite{li2019rethinking}&\underline{62.7}&\underline{77.0}&\underline{69.1}&56.4&\textbf{81.4}&66.7&47.4&47.6&47.5&-&-&-&36.3&\underline{42.8}&39.3\\
      Zhu \etal~\cite{zhu2019learning} &57.3&67.1&61.8&55.3&72.6&62.6&47.0&54.8&50.6&-&-&-&45.3&36.8&40.6\\
      SE-GZSL~\cite{kumar2018generalized}&56.3&67.8&61.5&58.3&68.1&62.8&41.5&53.3&46.7&-&-&-&40.9&30.5&34.9\\
      f-CLSWGAN~\cite{xian2018feature}&57.9&61.4&59.6&-&-&-&43.7&57.7&49.7&59.0&73.8&65.6&42.6&36.6&39.4\\
      cycle-CLSWGAN~\cite{felix2018multi}&56.9&64.0&60.2&-&-&-&45.7&61.0&52.3&59.2&72.5&65.1&49.4&33.6&40.0\\
      CADA-VAE~\cite{schonfeld2019generalized}&57.3&72.8&64.1&55.8&75.0&63.9&51.6&53.5&52.4&-&-&-&47.2&35.7&40.6\\
      f-VAEGAN-D2~\cite{xian2019f}&-&-&-&57.6&70.6&63.5&48.4&60.1&53.6&56.8&74.9&64.6&45.1&38.0&41.3\\
      LisGAN~\cite{li2019leveraging}&52.6&76.3&62.3&-&-&-&46.5&57.9&51.6&57.7&\underline{83.8}&68.3&42.9&37.8&40.2\\
      RFF-GZSL~\cite{han2020learning}&59.8&75.1&66.5&-&-&-&52.6&56.6&54.6&\underline{65.2}&78.2&71.1&45.7&38.6&41.9\\
      IZF~\cite{shen2020invertible}&61.3&\textbf{80.5}&\textbf{69.6}&60.6&77.5&\underline{68.0}&52.7&\textbf{68.0}&\underline{59.4}&-&-&-&\textbf{52.7}&\textbf{57.0}&\textbf{54.8}\\
      TF-VAEGAN~\cite{narayan2020latent}&-&-&-&59.8&75.1&66.6&52.8&64.7&58.1&62.5&\textbf{84.1}&\underline{71.7}&45.6&40.7&43.0\\
      \midrule
      \textbf{Our CE-GZSL} &\textbf{65.3}&73.4&\underline{69.1}&\textbf{63.1}&\underline{78.6}&\textbf{70.0}&\textbf{63.9}&\underline{66.8}&\textbf{65.3}&\textbf{69.0}&78.7&\textbf{73.5}&48.8&38.6&\underline{43.1}\\
      \bottomrule
    \end{tabular}
  }
  \vspace{-0.2cm}
\end{table*}

\paragraph{Implementation Details}
We implement our method with PyTorch. On all datasets, we set the
dimension of the embedding $h$ to 2,048, and set the dimension of the
non-linear projection's output $z$  to 512. The comparator network $F$
is a multi-layer perceptron (MLP) containing a hidden layer with
LeakyReLU activation. The comparator network $F$ takes as input the
concatenation of an embedding $h$ and a semantic descriptor $a$, and
outputs the relevance estimation between them. Our generator $G$ and
discriminator $D$ both contain a 4096-unit hidden layer with LeakyReLU
activation.
We use a random mini-batch size of 4,096 for AWA1 and AWA2, 2,048 for
CUB, 3,072 for FLO, and 1,024 for SUN in our method. In the
mini-batch, the instances from the same class are positive instances
to each other, while the instances from different classes are negative
instances to each other. The large batch size ensures a large number
of negative instances in our method.

\begin{table}[t]
  \caption{Results of conventional ZSL. The first six methods are
    early conventional ZSL methods and the following ten methods are
    recent proposed GZSL methods. The best results and the second best
    results are respectively marked in bold and underlined.}
  \centering
  \resizebox{0.9\linewidth}{!}
  {
    \label{tab:conventional}
    \begin{tabular}{l|c|c|c|c|c}
      \toprule
      Method&AWA1&AWA2&CUB&FLO&SUN\\
      \midrule
      LATEM~\cite{xian2016latent}&55.1&55.8&49.3&40.4&55.3\\
      DEVISE~\cite{frome2013devise}&54.2&59.7&52.0&45.9&56.5\\
      SJE~\cite{akata2015evaluation}&65.6&61.9&53.9&53.4&53.7\\
      ALE~\cite{akata2013label}&59.9&62.5&54.9&48.5&58.1\\
      ESZSL~\cite{romera2015embarrassingly}&58.2&58.6&53.9&51.0&54.5\\
      SYNC~\cite{changpinyo2016synthesized}&54.0&46.6&55.6&-&56.3\\
      \midrule
      DCN~\cite{liu2018generalized}&65.2&-&56.2&-&61.8\\
      SP-AEN~\cite{chen2018zero}&58.5&-&55.4&-&59.2\\
      cycle-CLSWGAN~\cite{felix2018multi}&66.3&-&58.4&70.1&60.0\\
      LFGAA~\cite{liu2019attribute}&-&68.1&\underline{67.6}&-&61.5\\
      DLFZRL~\cite{tong2019hierarchical}&\textbf{71.3}&70.3&61.8&-&61.3\\
      Zhu \etal~\cite{zhu2019learning}&69.3&70.4&58.5&-&61.5\\
      TCN~\cite{jiang2019transferable}&70.3&\underline{71.2}&59.5&-&61.5\\
      f-CLSWGAN~\cite{xian2018feature}&68.2&-&57.3&67.2&60.8\\
      f-VAEGAN-D2~\cite{xian2019f}&-&71.1&61.0&67.7&\underline{64.7}\\
      TF-VAEGAN~\cite{narayan2020latent}&-&\textbf{72.2}&64.9&\textbf{70.8}&\textbf{66.0}\\
      \midrule
      \textbf{Our CE-GZSL} &\underline{71.0}&70.4&\textbf{77.5}&\underline{70.6}&63.3\\
      \bottomrule
    \end{tabular}
  }
\end{table}

\begin{table*}[ht!]
  \caption{The effect of the hybrid GZSL framework. `Gen' denotes the
    feature generation model, `SE' denotes the semantic embedding
    model, and `+' denotes their hybrid combination. We evaluate these
    methods in three spaces: visual space (`V'), semantic space (`S'),
    and a new embedding space (`E').
  }
  \vspace{-0.1cm}
  \centering
  \resizebox{0.9\textwidth}{!}
  {
    \label{table:gen_embed}
    \begin{tabular}{lc|ccc|ccc|ccc|ccc|ccc}
      \toprule
      \multirow{2}*{Method}&\multirow{2}*{Space}&\multicolumn{3}{|c|}{AWA1}&\multicolumn{3}{c|}{AWA2}&\multicolumn{3}{c|}{CUB}&\multicolumn{3}{c|}{FLO}&\multicolumn{3}{c}{SUN}\\
                           &&${U}$&${S}$&${H}$&${U}$&${S}$&${H}$&${U}$&${S}$&${H}$&${U}$&${S}$&${H}$&${U}$&${S}$&${H}$\\
      \midrule
      Gen&V&53.0&67.7&59.5&56.9&61.6&59.2&54.1&59.4&56.6&57.5&75.5&65.3&43.0&\textbf{37.2}&39.9\\
      SE&S&21.8&55.7&31.3&21.1&59.9&31.2&36.3&44.2&39.9&24.0&62.6&34.7&19.0&27.1&22.4\\
      \midrule
      Gen+SE (basic)&S&50.5&62.5&55.9&50.6&64.3&56.6&52.2&59.3&55.5&53.2&\textbf{78.6}&63.4&35.1&23.3&28.0\\
      Gen+SE&E&\textbf{63.1}&\textbf{71.3}&\textbf{66.9}&\textbf{61.7}&\textbf{75.6}&\textbf{67.9}&\textbf{61.1}&\textbf{65.3}&\textbf{63.1}&\textbf{66.1}&72.2&\textbf{69.0}&\textbf{47.9}&36.1&\textbf{41.1}\\
      % \checkmark&\checkmark&&Gen+Semantic (softmax)&50.5&62.5&55.9&50.6&64.3&56.6&52.2&59.3&55.5&53.2&78.6&63.4\\
      % \checkmark&\checkmark&N&softmax&63.1&71.3&66.9&61.7&75.6&67.9&61.1&65.3&63.1&66.1&72.2&69.0\\
      \bottomrule
    \end{tabular}
  }
  \vspace{-0.1cm}
\end{table*}

\begin{table*}[ht!]
  \caption{The effect of different embedding models (E-M) and
    different spaces in the hybrid GZSL framework. All the methods
    here are combined with the feature generation model. `SE' denotes
    the semantic embedding model and `CE' denotes our contrastive
    embedding model. We evaluate the embedding models in two embedding
    spaces: semantic descriptor space (S) and the new embedding space
    (E).}
  \vspace{-0.1cm}
  \centering
  \resizebox{0.9\textwidth}{!}
  {
    \label{table:embedding_space}
    \begin{tabular}{c|l|ccc|ccc|ccc|ccc|ccc}
      \toprule
      \multirow{2}*{Space}&\multirow{2}*{E-M}&\multicolumn{3}{|c|}{AWA1}&\multicolumn{3}{c|}{AWA2}&\multicolumn{3}{c|}{CUB}&\multicolumn{3}{c|}{FLO}&\multicolumn{3}{c}{SUN}\\
       &&${U}$&${S}$&${H}$&${U}$&${S}$&${H}$&${U}$&${S}$&${H}$&${U}$&${S}$&${H}$&${U}$&${S}$&${H}$\\
      \midrule
      V&None&53.0&67.7&59.5&56.9&61.6&59.2&54.1&59.4&56.6&57.5&75.5&65.3&43.0&37.2&39.9\\
      \midrule
      \multirow{2}*{S}
       &SE (basic)&50.5&62.5&55.9&50.6&64.3&56.6&52.2&59.3&55.5&53.2&78.6&63.4&35.1&23.3&28.0\\
       &CE&55.0&65.9&59.9&55.8&70.7&62.4&61.5&\textbf{67.4}&64.3&56.1&\textbf{78.9}&65.5&37.6&30.4&33.6\\
      \midrule
      \multirow{2}*{E}
       &SE&63.1&71.3&66.9&61.7&75.6&67.9&61.1&65.3&63.1&66.1&72.2&69.0&47.9&36.1&41.1\\
       &CE (\textbf{Our CE-GZSL})&\textbf{65.3}&\textbf{73.4}&\textbf{69.1}&\textbf{63.1}&\textbf{78.6}&\textbf{70.0}&\textbf{63.9}&66.8&\textbf{65.3}&\textbf{69.0}&78.7&\textbf{73.5}&\textbf{48.8}&\textbf{38.6}&\textbf{43.1}\\
      \bottomrule
    \end{tabular}
  }
  \vspace{-0.1cm}
\end{table*}

\begin{table*}[ht!]
  \caption{Evaluation of each part of our contrastive embedding (CE)
    model in the hybrid GZSL framework. `\textbf{Our CE-GZSL}' denotes
    the whole CE model.}
  \vspace{-0.1cm}
  \centering
  \resizebox{0.9\textwidth}{!}
  {
    \label{table:comp_ablation}
    \begin{tabular}{l|ccc|ccc|ccc|ccc|ccc}
      \toprule
      \multirow{2}*{Method} &\multicolumn{3}{|c|}{AWA1}&\multicolumn{3}{c|}{AWA2}&\multicolumn{3}{c|}{CUB}&\multicolumn{3}{c|}{FLO}&\multicolumn{3}{c}{SUN}\\
                            &${U}$&${S}$&${H}$&${U}$&${S}$&${H}$&${U}$&${S}$&${H}$&${U}$&${S}$&${H}$&${U}$&${S}$&${H}$\\
      \midrule
     $V(G,D)+\mathcal{L}_{ce}^{ins}(G,E,H)$&64.7&71.3&67.8&\textbf{64.4}&72.3&68.1&58.8&66.5&62.4&62.9&77.3&69.4&49.0&32.0&38.7\\
      $V(G,D)+\mathcal{L}_{ce}^{cls}(G,E,F)$ &63.6&72.0&67.5&61.2&\textbf{79.3}&69.1&62.7&63.3&63.0&66.0&\textbf{79.7}&72.2&\textbf{49.1}&37.4&42.4\\
      \midrule
      \textbf{Our CE-GZSL}&\textbf{65.3}&\textbf{73.4}&\textbf{69.1}&63.1&78.6&\textbf{70.0}&\textbf{63.9}&\textbf{66.8}&\textbf{65.3}&\textbf{69.0}&78.7&\textbf{73.5}&48.8&\textbf{38.6}&\textbf{43.1}\\
      \bottomrule
    \end{tabular}
  }
  \vspace{-0.1cm}
\end{table*}

\subsection{Comparison with SOTA}
In Table~\ref{table:comp_sota}, we compare our CE-GZSL method with the
state-of-the-art GZSL methods. Our method achieves the best $U$ on
four datasets and achieves the best $H$ on AWA2, CUB, and
FLO. Notably, on CUB, our CE-GZSL is the first one that obtains
the performances $>60.0$ on $U$ and $H$ among the state-of-the-art
GZSL methods. Especially, our hybrid GZSL method integrating with the
simplest generative model still achieves competitive results compared
with IZF~\cite{shen2020invertible}, which is based on the most advanced
generative model in GZSL. Our CE-GZSL achieves the second best $H$ on
AWA1 and SUN, and is only lower than IZF~\cite{shen2020invertible},
and on the other three datasets our CE-GZSL outperforms
IZF~\cite{shen2020invertible} by a large margin. In
Table~\ref{tab:conventional}, we report the results of our CE-GZSL
under the conventional ZSL scenario. We compare our method with 
sixteen methods, in which six of them are traditional methods and ten
of them are the recent methods. Our method is still competitive in
conventional ZSL. Our method performs the best on CUB and the second
best on AWA1 and FLO in the conventional ZSL scenario. Specifically,
on CUB, our method also achieves an excellent performance, and our
CE-GZSL is the only method that can achieve the performance $>70.0$
under conventional ZSL among the ten recent methods.

\subsection{Component Analysis}
In Table~\ref{table:gen_embed}, we illustrate the effectiveness of the
hybrid strategy for GZSL. First, we respectively evaluate the
performances of the single feature generation model (Gen) and the  
single semantic embedding model (SE). We evaluate them in their
original space: visual space (V) for `Gen' and semantic space (S) for
`SE'. `Gen+SE (basic)' denotes that we simply combine the feature
generation model with the semantic embedding model and learn a softmax
classifier in semantic space, corresponding to the basic hybrid GZSL
approach defined in Eq.~\ref{eq:se_gen_loss}.
Moreover, we introduce a new embedding space (E) in the hybrid GZSL
method, which leads to the increased performance. The results show
that the hybrid GZSL strategy is effective, and the new embedding
space is better than the semantic space.

In Table~\ref{table:embedding_space}, we investigate the effect of
different spaces and different embedding models in the hybrid GZSL
framework. We integrate the feature generation model with two
different embedding models: semantic embedding (SE) (i.e. the ranking
loss method) and our contrastive embedding (CE). And we evaluate the
semantic descriptor space (S) and the new embedding space (E), in
which we conduct the final GZSL classification. Firstly, we evaluate
the same embedding model on different embedding spaces: the results of
`SE' on the new embedding space performs much better than `SE (basic)'
on the semantic space; and `CE (\textbf{Our CE-GZSL})' on the new
embedding space also performs better than `CE' on the semantic
descriptor space. This demonstrates that the new embedding space is
much more effective than the original semantic space in our hybrid
framework. Afterward, we compare the results on the same embedding
space but using different embedding models: `SE' corresponds to the
ranking loss form in Eq.~\ref{eq:L_se_sync} and `CE' corresponds to
contrastive form in Eq.~\ref{eq:l_ce_cls}. Our proposed `CE' can
always outperform `SE', no matter in the semantic descriptor space or
in the new embedding space. This illustrates that our contrastive
embedding (CE) benefits from the instance-wise supervision which is
neglected in the traditional semantic embedding (SE).

Moreover, in Table~\ref{table:comp_ablation}, we respectively evaluate
the instance-level supervision and the class-level supervision in our
contrastive embedding model. Concretely, to evaluate the
instance-level supervision, we remove the class-level supervision
$\mathcal{L}_{ce}^{cls}(G,E,F)$ in Eq.~\ref{eq:ce_gen_loss} and only
optimize $V(G,D)+\mathcal{L}_{ce}^{ins}(G,E,H)$ to learn our
contrastive embedding model. In the same way, we evaluate the
class-level supervision by optimizing
$V(G,D)+\mathcal{L}_{ce}^{cls}(G,E,F)$. As shown in
Table~\ref{table:comp_ablation}, when using either the instance-level
CE or the class-level CE, our result is still competitive compared
with the state-of-the-art GZSL methods. When considering both the
instance-level supervision and the class-level supervisions, our
method achieves the improvements on $U$ and $S$, leading to the
better $H$ results. This means that our method benefits from the
combination of the instance-level supervision and the class-level
supervision.

\begin{figure*}[ht!]
  \centering
  \begin{subfigure}[b]{0.19\textwidth}
    \centering
    \includegraphics[width=\textwidth]{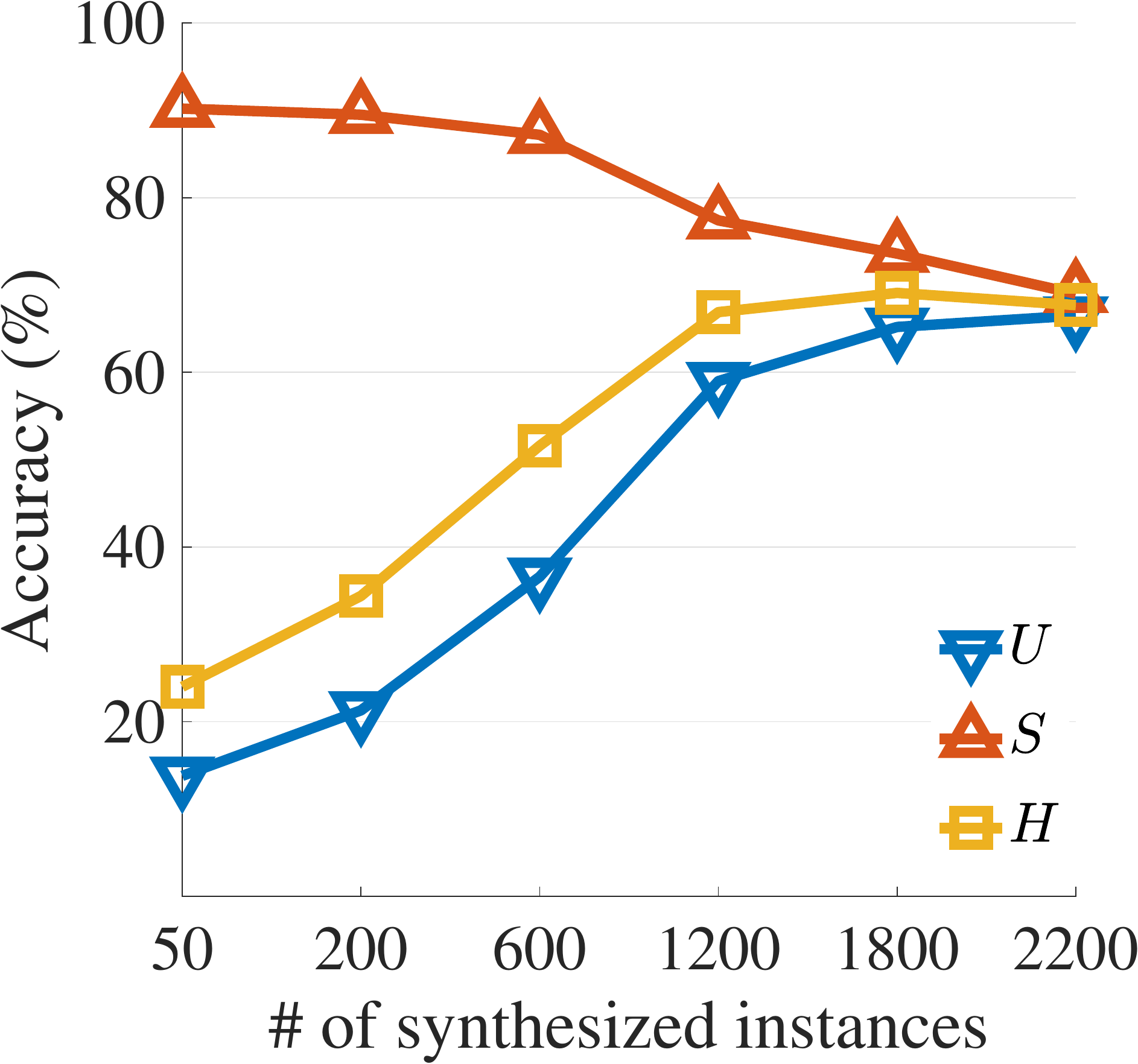}
    \caption{\footnotesize{AWA1}}
    \label{fig:AWA1_syn}
  \end{subfigure}
  \begin{subfigure}[b]{0.19\textwidth}
    \centering
    \includegraphics[width=\textwidth]{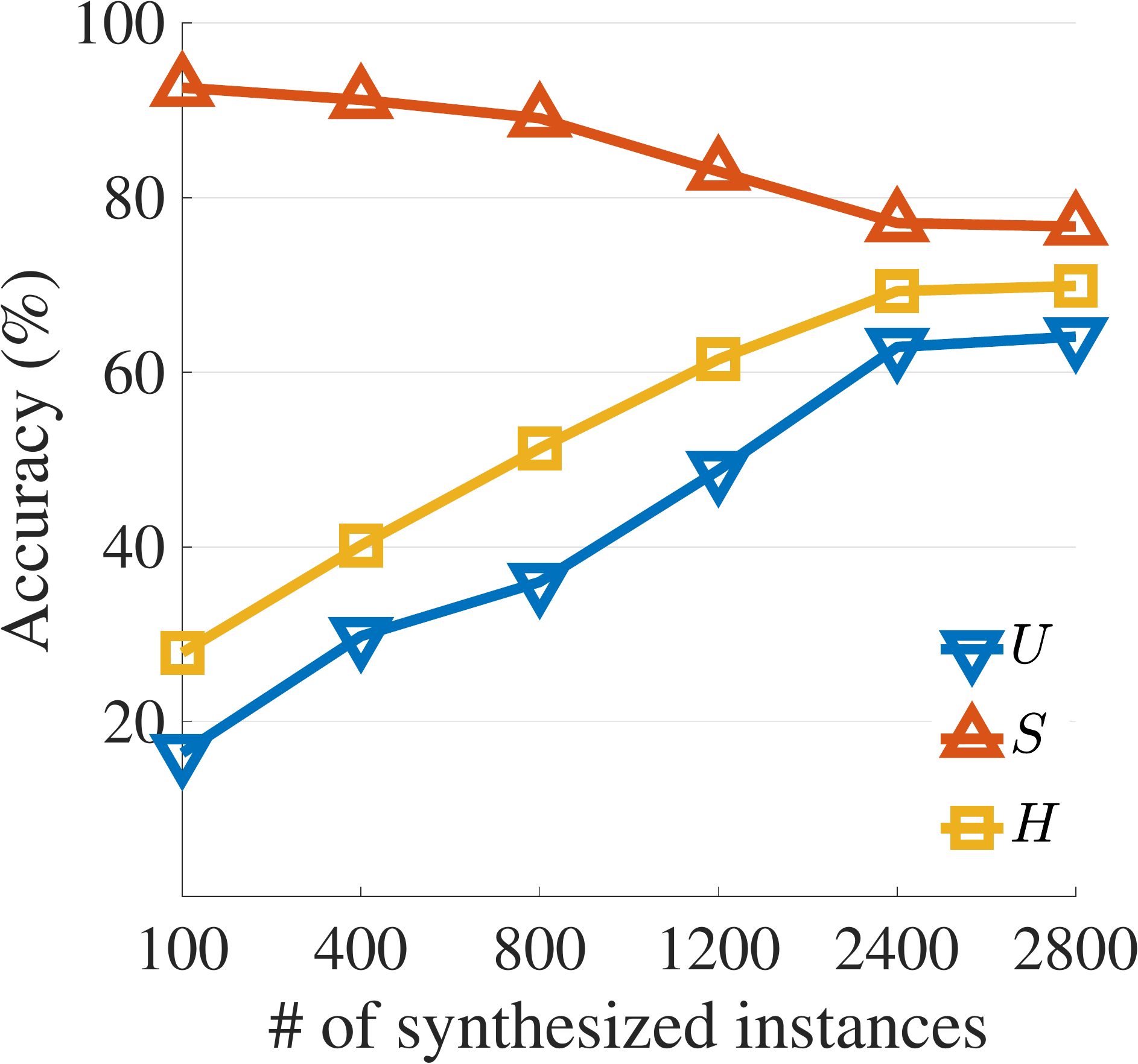}
    \caption{\footnotesize{AWA2}}
    \label{fig:AWA2_syn}
  \end{subfigure}
  \begin{subfigure}[b]{0.19\textwidth}
    \centering
    \includegraphics[width=\textwidth]{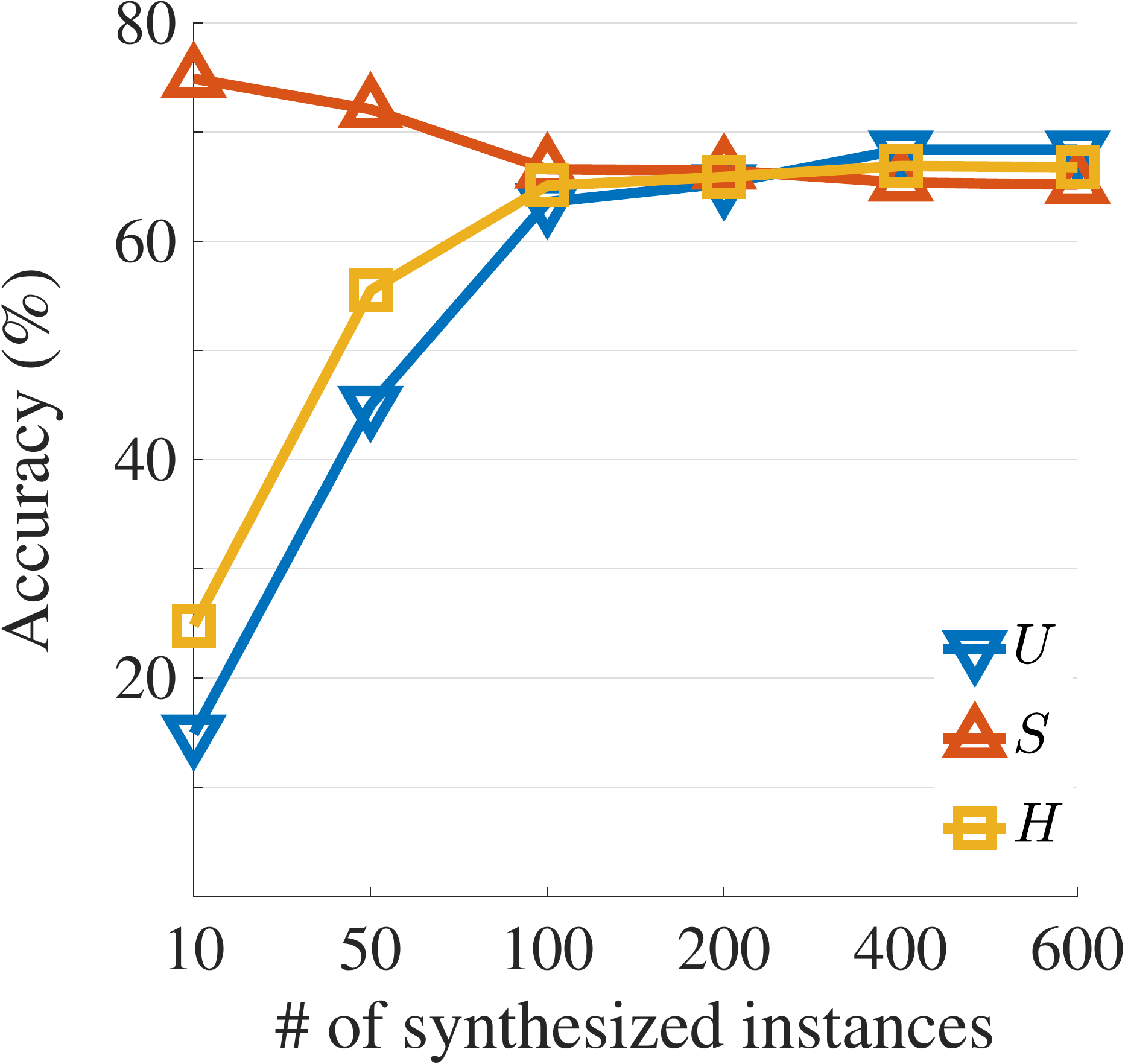}
    \caption{\footnotesize{CUB}}
    \label{fig:CUB_syn}
  \end{subfigure}
  \begin{subfigure}[b]{0.19\textwidth}
    \centering
    \includegraphics[width=\textwidth]{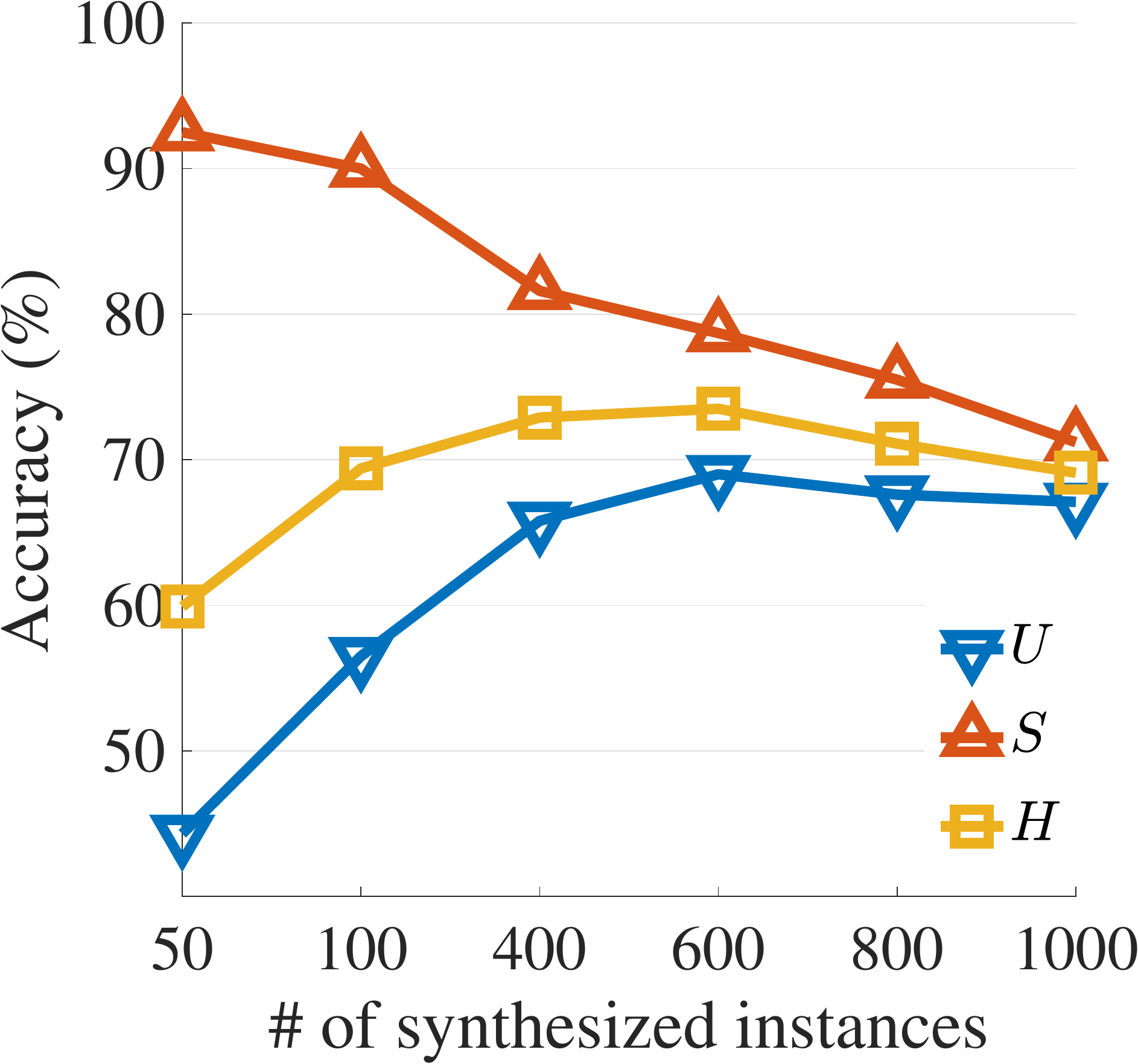}
    \caption{\footnotesize{FLO}}
    \label{fig:FLO_syn}
  \end{subfigure}
  \begin{subfigure}[b]{0.19\textwidth}
	\centering
	\includegraphics[width=\textwidth]{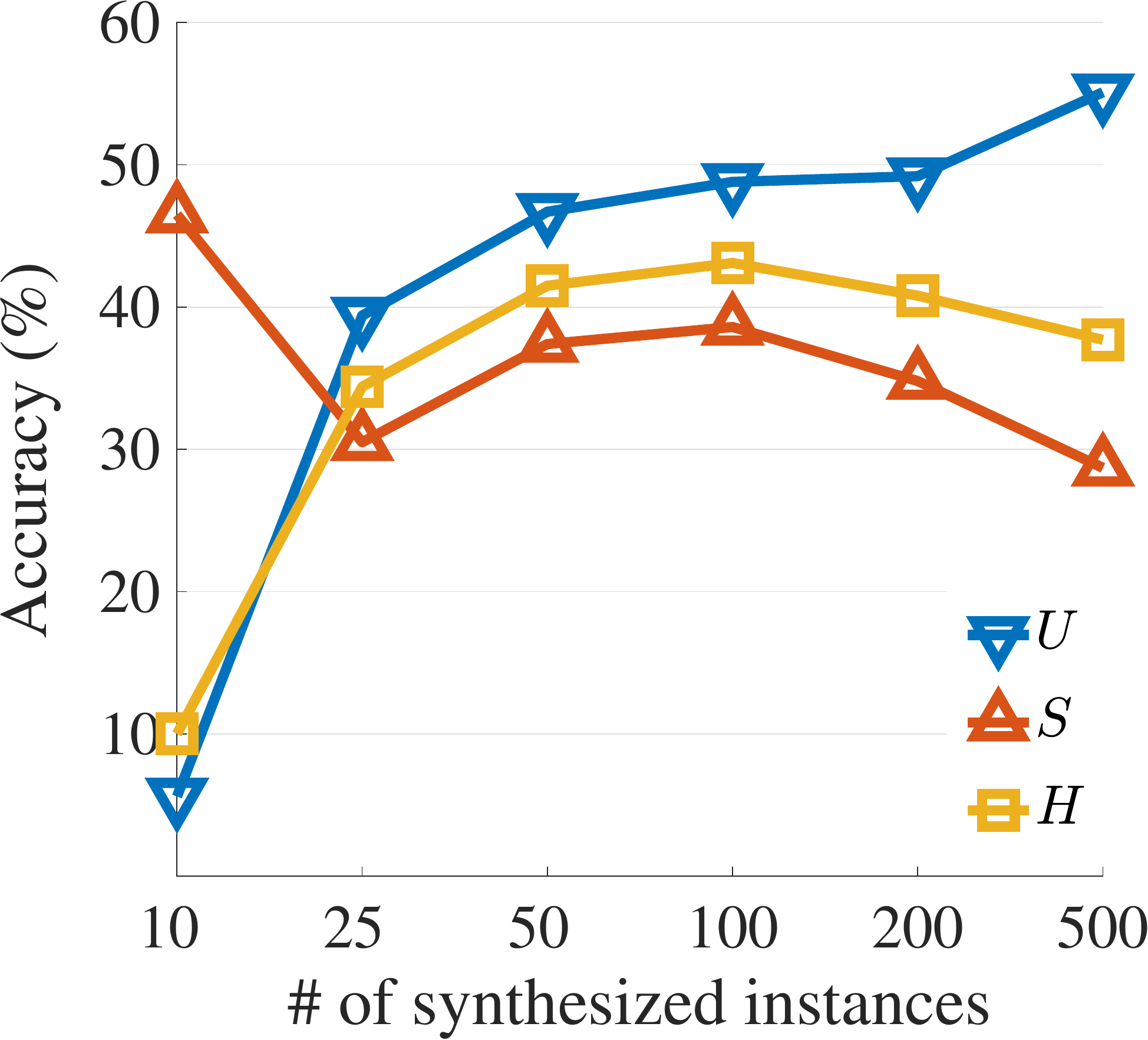}
	\caption{\footnotesize{SUN}}
	\label{fig:SUN_syn}
  \end{subfigure}
  \vspace{-0.1cm}
  \caption{The GZSL results with respect to different numbers of the
    synthesized samples for each unseen class.}
  \vspace{-0.1cm}
  \label{fig:syn_effect}
\end{figure*}

\begin{figure*}[ht!]
  \centering
  \begin{subfigure}[b]{0.19\textwidth}
    \centering
    \includegraphics[width=\textwidth]{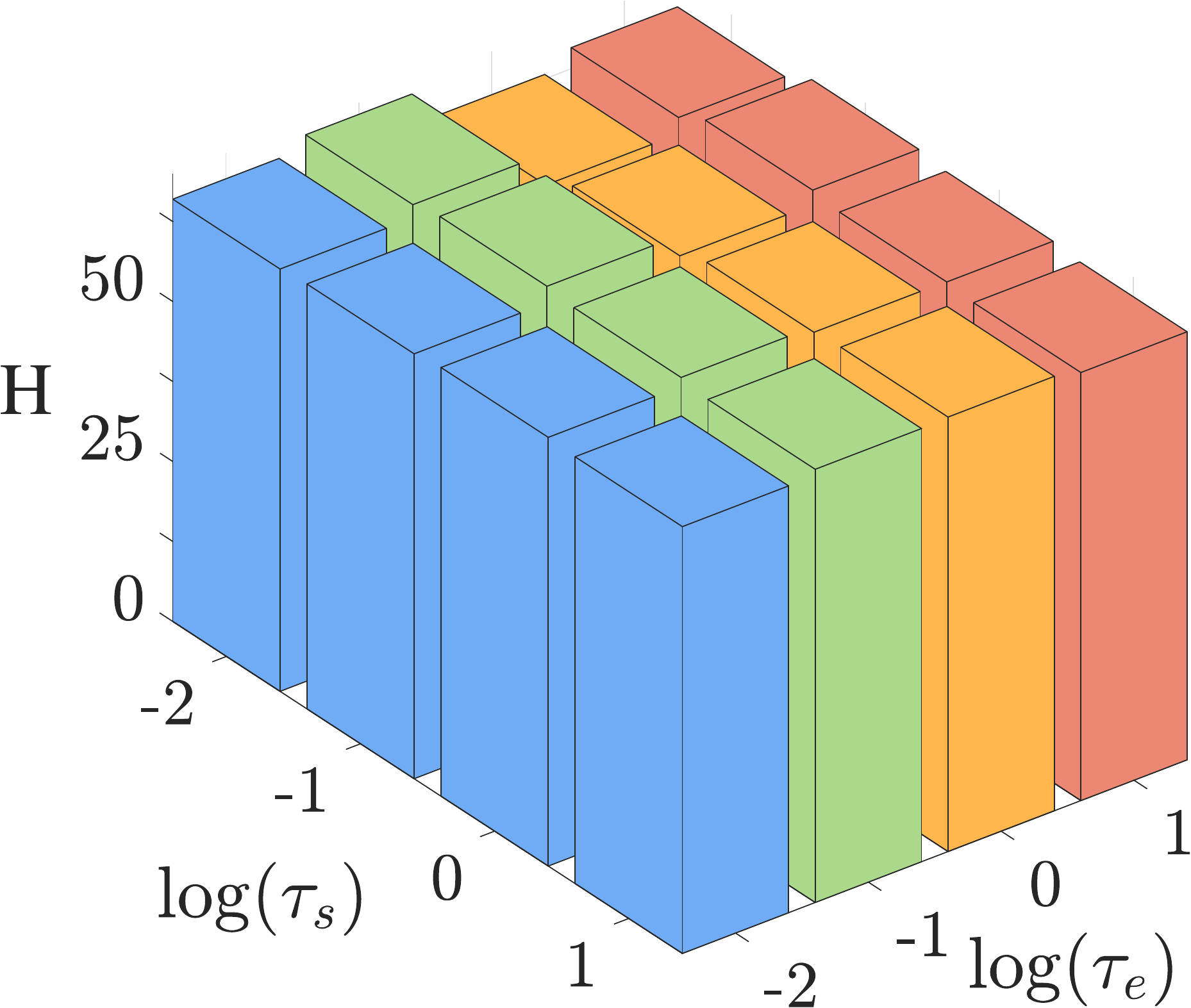}
    \caption{\footnotesize{AWA1}}
    \label{fig:AWA1_temp}
  \end{subfigure}
  \begin{subfigure}[b]{0.19\textwidth}
    \centering
    \includegraphics[width=\textwidth]{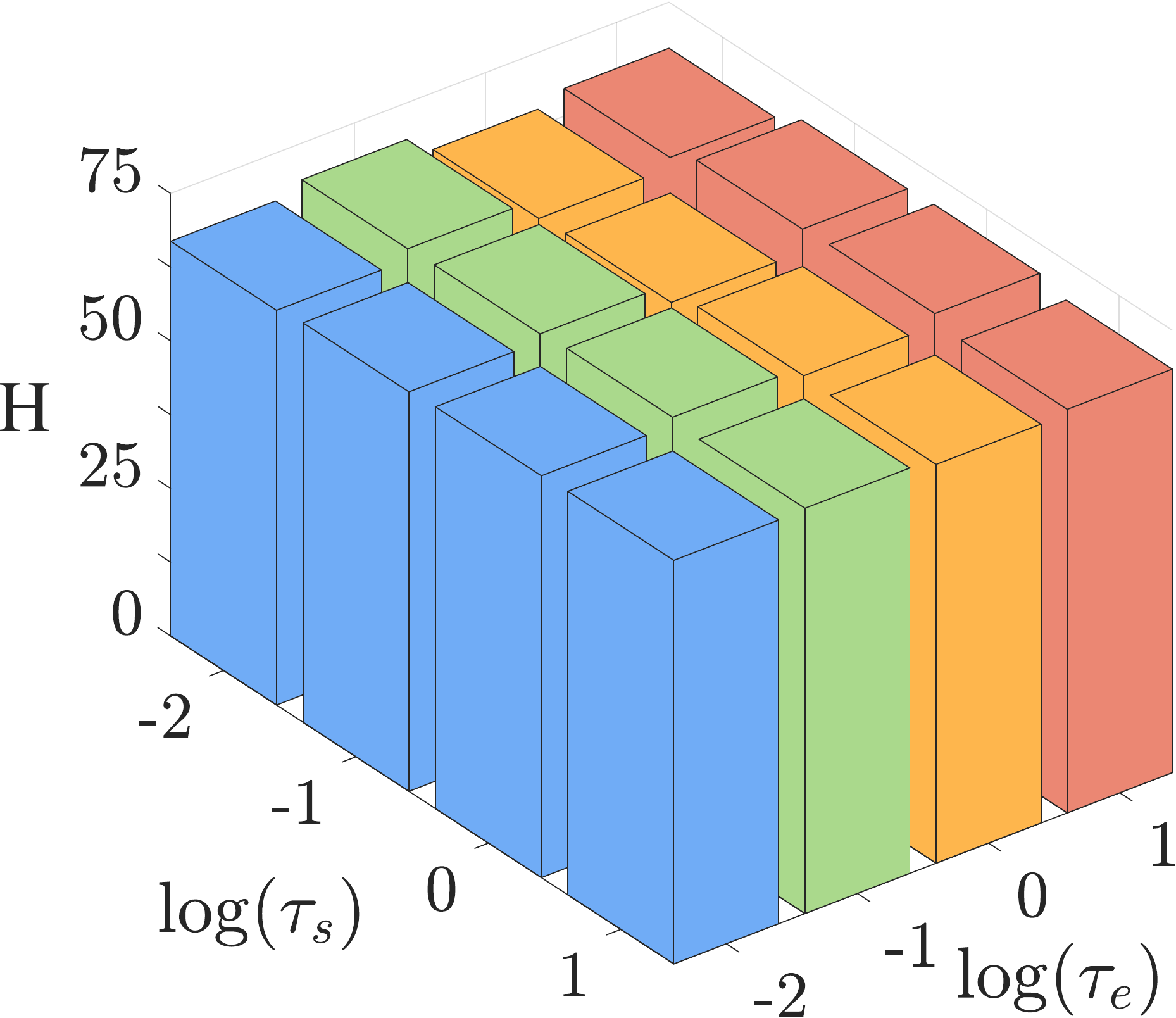}
    \caption{\footnotesize{AWA2}}
    \label{fig:AWA2_temp}
  \end{subfigure}
  \begin{subfigure}[b]{0.19\textwidth}
    \centering
    \includegraphics[width=\textwidth]{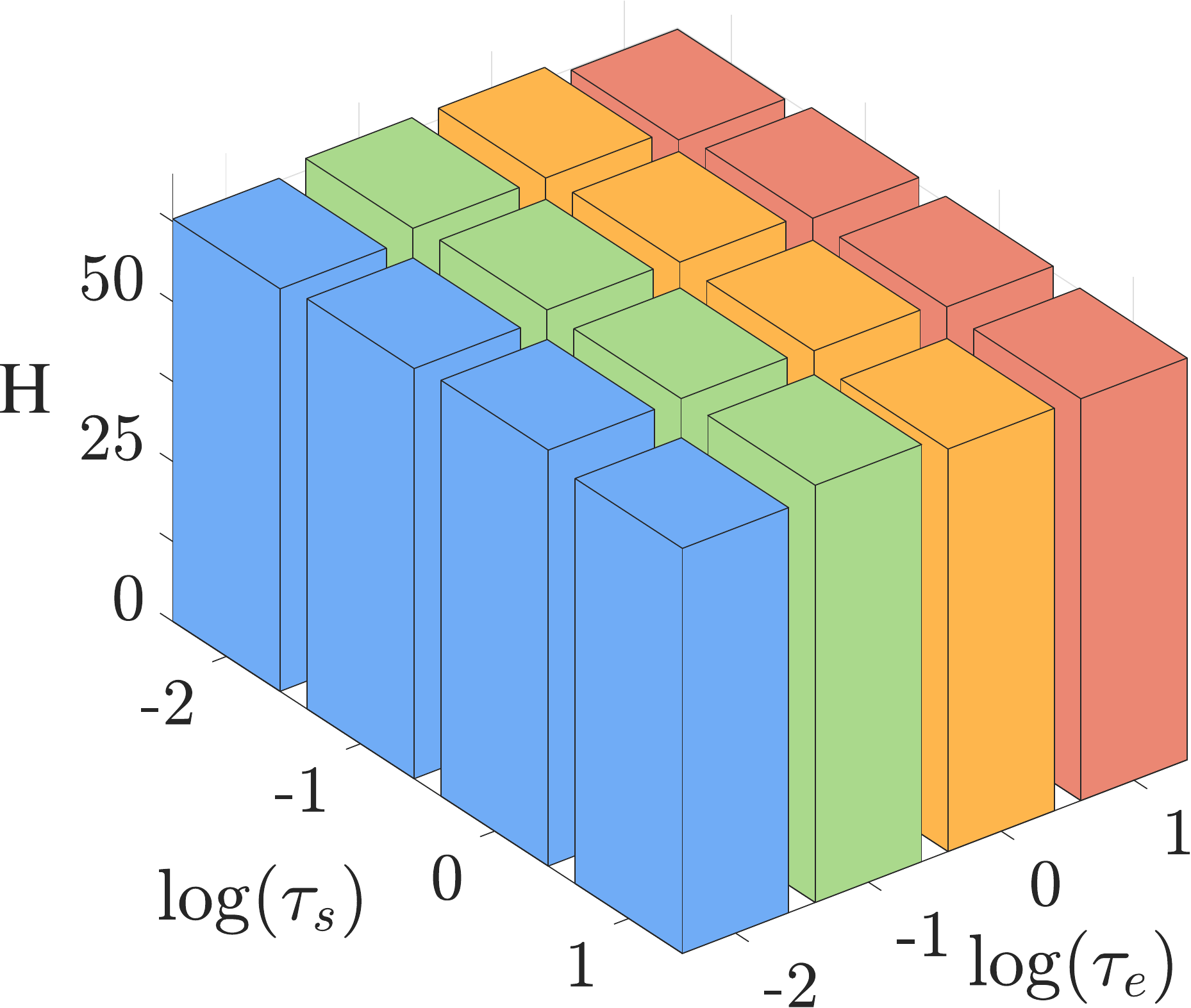}
    \caption{\footnotesize{CUB}}
    \label{fig:CUB_temp}
  \end{subfigure}
  \begin{subfigure}[b]{0.19\textwidth}
    \centering
    \includegraphics[width=\textwidth]{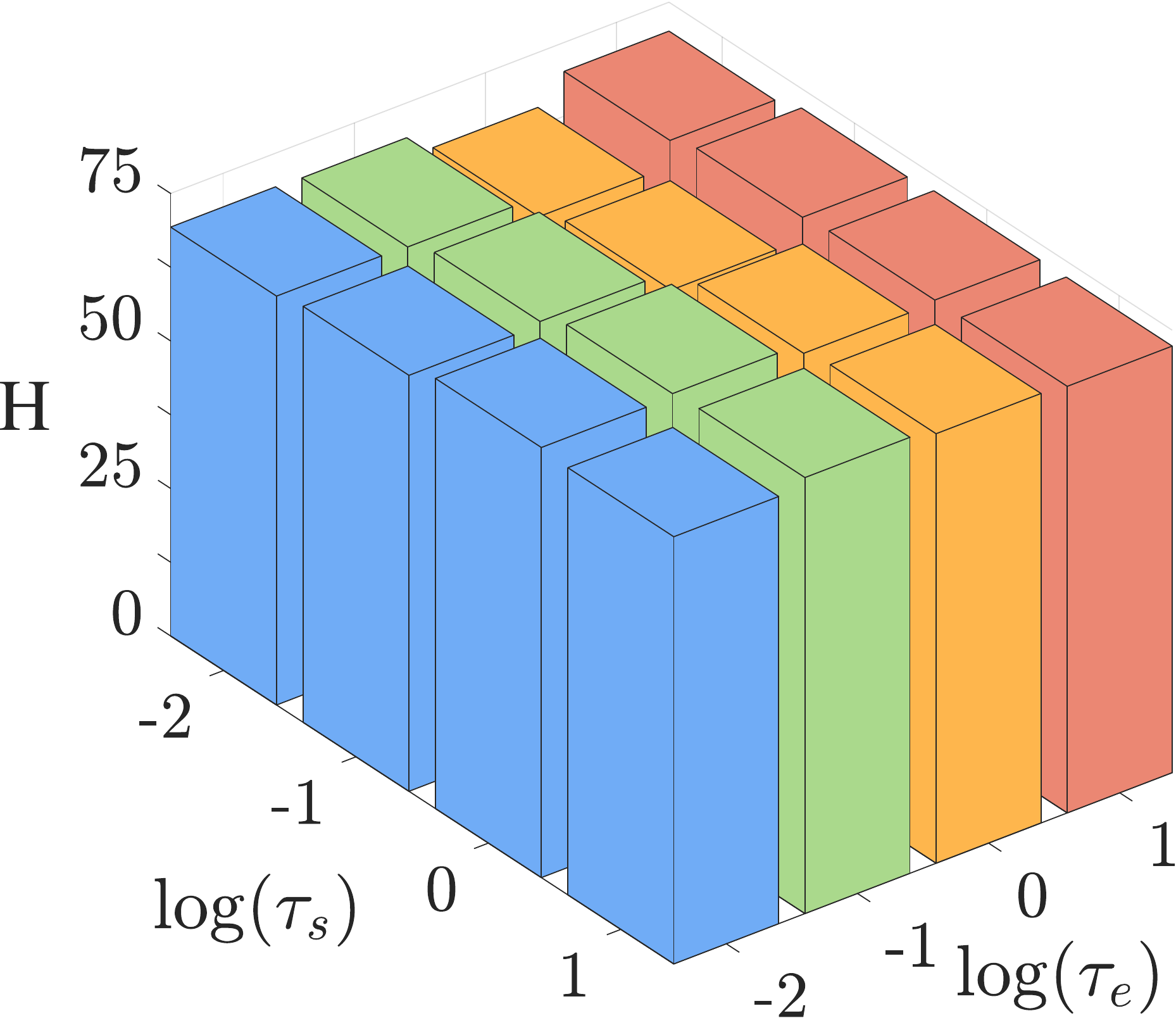}
    \caption{\footnotesize{FLO}}
    \label{fig:FLO_temp}
  \end{subfigure}
  \begin{subfigure}[b]{0.19\textwidth}
    \centering
    \includegraphics[width=\textwidth]{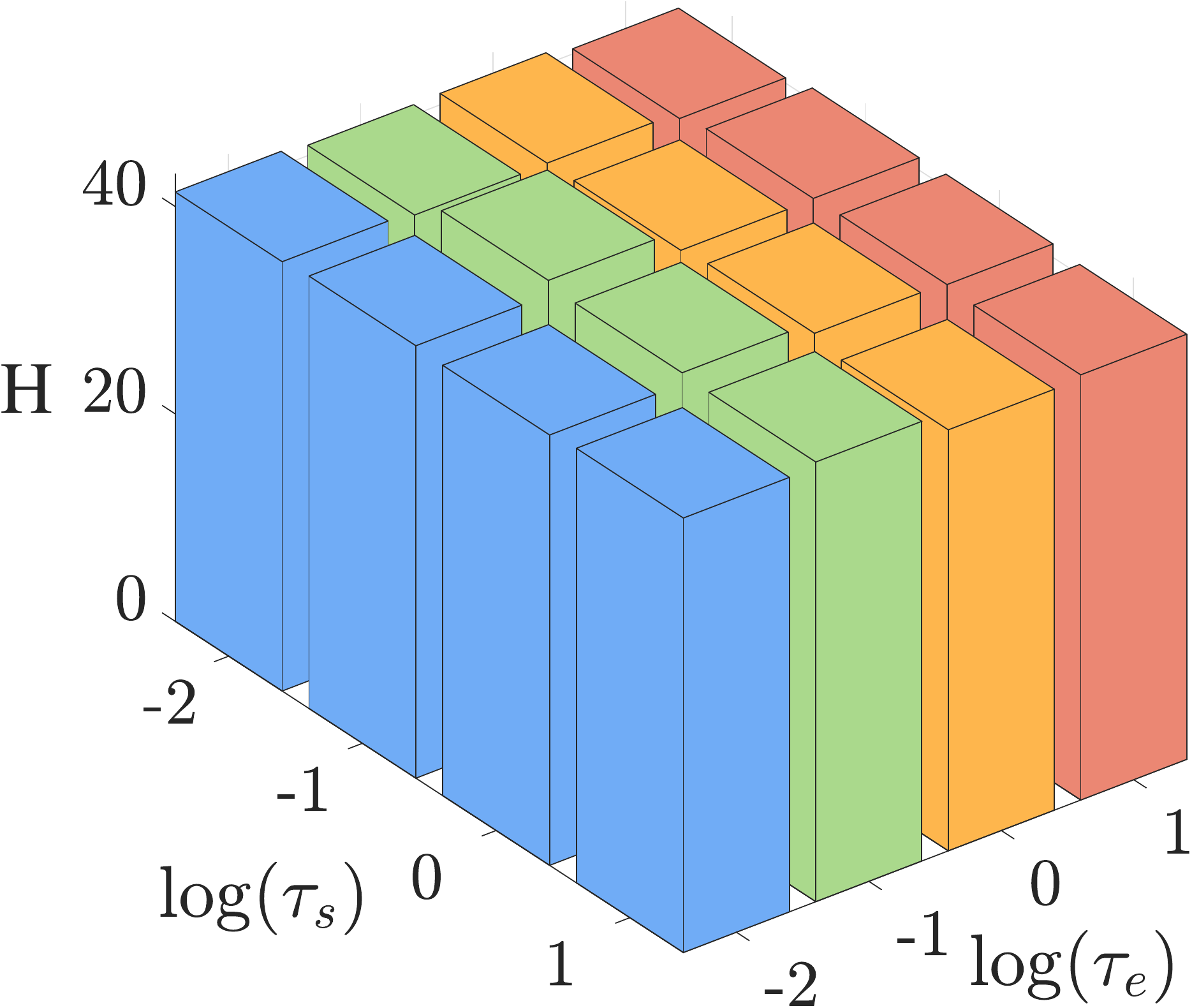}
    \caption{\footnotesize{SUN}}
    \label{fig:SUN_temp}
  \end{subfigure}
  \vspace{-0.1cm}
  \caption{The results of harmonic mean $H$ in GZSL with respect to
    different temperature parameters $\tau_e$ and $\tau_s$.}
  \vspace{-0.1cm}
  \label{fig:temp_effect}
\end{figure*}

\begin{table*}[ht!]
	\caption{Our CE-GZSL results with respect to different dimensions of
		the embedding $h$. $U$ and $S$ are the Top-1 accuracies tested
		on unseen classes and seen classes, respectively, in GZSL. $H$ is
		the harmonic mean of $U$ and $S$. }
	\centering
	\resizebox{0.9\textwidth}{!}
	{
		\label{table:dim}
		\begin{tabular}{l|ccc|ccc|ccc|ccc|ccc}
			\toprule
			\multirow{2}*{Dimension}&\multicolumn{3}{|c|}{AWA1}&\multicolumn{3}{c|}{AWA2}&\multicolumn{3}{c|}{CUB}&\multicolumn{3}{c}{FLO}&\multicolumn{3}{c}{SUN}\\
			&${U}$&${S}$&${H}$&${U}$&${S}$&${H}$&${U}$&${S}$&${H}$&${U}$&${S}$&${H}$&${U}$&${S}$&${H}$\\
			\midrule
			256&52.9&53.9&53.3&56.6&71.8&63.3&62.5&51.5&62.0&63.0&47.2&53.9&46.7&28.9&35.7\\
			512&65.0&55.7&60.0&60.7&75.4&67.3&62.2&66.1&64.1&68.1&58.0&62.6&\textbf{50.1}&33.6&40.2\\
			1,024&\textbf{65.7}&65.1&65.4&61.9&77.9&69.0&62.2&66.5&64.3&63.2&78.0&69.8&48.7&36.7&41.9\\
			2,048&64.9&73.9&\textbf{69.0}&63.1&\textbf{78.6}&\textbf{70.0}&\textbf{63.9}&66.8&\textbf{65.3}&\textbf{69.0}&78.7&73.5&48.8&38.6&\textbf{43.1}\\
			4,096&62.9&\textbf{76.4}&\textbf{69.0}&\textbf{66.0}&74.5&\textbf{70.0}&63.0&\textbf{67.6}&65.2&68.5&\textbf{80.3}&\textbf{73.9}&45.6&\textbf{39.0}&42.1\\
			\bottomrule
		\end{tabular}
	}
\end{table*}

\subsection{Hyper-Parameter Analysis}
We evaluate the effect of different numbers of synthesized instances
per unseen classes as shown in Figure~\ref{fig:syn_effect}. The
performances on five datasets increase along with the number of
synthesized examples, which shows the data-imbalance problem has been
relieved by the generation model in our hybrid GZSL framework. Our
method achieves the best results on AWA1, AWA2, CUB, FLO, and SUN when we
synthesize 1,800, 2,400, 300, 600, and 100 examples per unseen classes,
respectively.

\begin{table}[t!]
	\caption{The effect of different numbers of positive and negative
		samples in a mini-batch on AWA1. `$P$' and `$K$ denote the numbers
		of positive examples and negative examples in the mini-batch,
		respectively.}
	\vspace{-0.2cm}
	\centering
	\resizebox{0.75\linewidth}{!}
	{
		\label{table:pos_neg}
		\begin{tabular}{cc|ccc}
			\toprule
			$P$ &$K$&${U}$&${S}$&${H}$\\
			\midrule
			1&50&57.5&74.6&64.9\\
			1&100&59.0&73.2&65.3\\
			1&500&63.1&70.6&66.7\\
			1&1,000&61.8&70.9&66.0\\
			1&2,000&60.9&73.5&66.6\\
			1&4,000&61.5&\textbf{74.7}&67.4\\
			\midrule
			30&50&61.8&72.6&66.8\\
			30&100&61.9&73.1&67.0\\
			30&500&64.2&72.3&68.0\\
			30&1,000&63.5&72.8&67.9\\
			30&2,000&63.7&73.4&68.2\\
			30&4,000&62.4&73.0&67.3\\
			\midrule
			\multicolumn{2}{c|}{random batch (4,096)}&\textbf{65.3}&73.4&\textbf{69.1}\\
			\bottomrule
		\end{tabular}
	}
	\vspace{-0.2cm}
\end{table}

Next, we evaluate the influence of the temperature parameters, $\tau_e$ and
$\tau_s$, in the contrastive embedding model. We cross-validate
$\tau_e$ and $\tau_s$ in $[0.01, 0.1, 1.0, 10.0]$ and plot the $H$
values with respect to different $\tau_e$ and $\tau_s$, as shown in
Figure~\ref{fig:temp_effect}. With the different $\tau_e$ and $\tau_s$
values, the $H$ results on different datasets change slightly,
indicating that our method is robust to the temperature parameters. On
AWA1, CUB and SUN, our method achieves the best results when
$\tau_e=0.1$ and $\tau_s=0.1$. On AWA2, our method achieves the best
result when $\tau_e=10.0$ and $\tau_s=1.0$. On FLO, our method
achieves the best result when $\tau_e=0.1$ and $\tau_s=1.0$.

In Table~\ref{table:dim}, we report the results of our hybrid GZSL
with contrastive embedding (CE-GZSL) with respect to different
dimensions of the embedding $h$.
On each of the datasets, as the dimension of the embedding $h$
grows, the performance of our CE-GZSL improves significantly.
However, a high dimensional embedding space will inevitably increase
the computational burden.
Thus, in our experiments, we set the dimension of the embedding $h$
to 2,048 in order to achieve a trade-off between performance and
computational cost.

We further evaluate the effect of the numbers of positive and negative
examples in the mini-batch. In a mini-batch, we sample $P$ positive
examples and $K$ negative examples for a given example. We report the
results on AWA1 in Table~\ref{table:pos_neg}. We can observe that our
method benefits from more positive examples and more negative
examples. We find that using a large random batch (4,096) without a
hand-crafted designed sampling strategy leads to the best results. The
reason is that a large batch will contain enough positive examples and
negative examples.

%-------------------------------------------------------
\section{Conclusion}
In this paper, we have proposed a hybrid GZSL framework, integrating
an embedding model and a generation model. The proposed hybrid GZSL
framework maps the real and synthetic visual samples into an embedding
space, where we can train a supervised recognition model as the final
GZSL classifier. Specifically, we have proposed a contrastive
embedding model in our hybrid GZSL framework. Our contrastive
embedding model can leverage not only the class-wise supervision but
also the instance-wise supervision. The latter is usually neglected in
existing GZSL researches. The experiments show that our hybrid GZSL
framework with contrastive embedding (CE-GZSL) has achieved the
state-of-the-arts on three benchmark datasets and achieved the
second-best on two datasets.

%\section*{Acknowledgment}
%This work was supported by the National Science Foundation of China
%(Grant No. U1713208 and 61876085) and the China Postdoctoral Science
%Foundation (Grant No. 2017M621748, 2020M681606 and 2019T120430).

{\small
\bibliographystyle{ieee_fullname}
\bibliography{egbib}
}

\end{document}